\documentclass[letterpaper, 10 pt,journal]{IEEEtran}
\usepackage{amsmath,amsfonts}
\usepackage{amsthm}
\usepackage[caption=false,font=normalsize,labelfont=sf,textfont=sf]{subfig}
\usepackage{graphicx}
\usepackage{hyperref}
\usepackage{multirow}
\usepackage{physics}
\usepackage{siunitx}
\usepackage{stfloats}
\usepackage{tabularx}
\usepackage{textcomp}
\usepackage{todonotes}
\usepackage{url}
\usepackage{verbatim}
\usepackage{wrapfig}
\usepackage{xcolor,soul}
\definecolor{lightblue}{rgb}{.8,.95,1}
\sethlcolor{lightblue}
\usepackage{xr}

\graphicspath{{./resources/}{./resources/tmp/}}
\DeclareGraphicsExtensions{.pdf,.svg,.png}

\begin{document}
\IEEEoverridecommandlockouts                              
	\title{Direct Rotor Thrust Sensing and Feedback Control for Disturbance Rejection of Multirotors Using Load-cells}
	\author{Peter B{\"o}hm$^{1}$ \textit{IEEE Member}, Michael Br{\"u}nig$^{2}$  \textit{Senior IEEE Member}, Peyman Moghadam$^{3}$ \textit{Senior IEEE Member}, and Pauline E.I. Pounds$^{4}$ \textit{IEEE Member}%
	\thanks{*This work was in part supported by the Advance Queensland Trusted Autonomous Systems Defence CRC Fellowship.}
	\thanks{$^{1}$Peter B{\"o}hm is a postdoctoral fellow,
		{\tt\small p.bohm@uq.edu.au}}
	\thanks{$^{2}$Michael Br{\"u}nig is a Professor and Head of School of EECS,
		{\tt\small m.bruenig@uq.edu.au}}
	\thanks{$^{3}$Peyman Moghadam is with CSIRO Robotics, CSIRO, Australia.
        {\tt\small peyman.moghadam@csiro.au}}%
	\thanks{$^{4}$Pauline E. I. Pounds is a Professor in Mechatronics, all at the University of Queensland, Queensland, Australia
		{\tt\small pauline.pounds@uq.edu.au}}}



\maketitle
\setcounter{footnote}{3}

\begin{abstract}
Gust disturbances, dynamic vertical inflow and ground effect are key adverse aerodynamic phenomena that induce variations in the forces acting on a multirotor and complicate its flight control.  Miniature rotorcraft typically rely on simplified modelling of such effects to compute adjustments in thrust to counteract these forces. In the most basic case, disturbance force estimations are derived from the aircraft's motion and the generated thrust is assumed to exactly match that requested by the controller. However, such systems rely on the aircraft's trajectory to be affected before disturbances can be sensed and compensated. Numerous approaches presented over the last 15-20 years aim to reject external disturbances more quickly, but challenges remain.

This paper presents a new approach in this category by measuring the instantaneous force of the rotors directly at the point of generation using load-cells and implementing high-speed control to accurately track the desired thrust. Measurements from load-cells were previously considered too noisy to provide meaningful input, but the experiments presented in the paper using purpose-built hardware from low-cost commodity components in single- and dual rotor see-saw models and a flying aircraft demonstrate both the feasibility and the effectiveness of the approach in the presence of complex aerodynamic phenomena.
\end{abstract}

\vspace{-10px}

\section{Introduction}\label{sec:intro}

%

Gust disturbances, vertical rotor inflow (eg. vortex ring state) and ground effect have long been complicating factors in the flight control of rotorcraft drones.  Compensating for such influences is the quintessential multirotor control task: computing rotor thrusts that reject disturbances, and steer the aircraft on a desired trajectory.  Rotorcraft dynamics are well understood, and control is typically based on simplified linear models for hovering in still air, with rotor thrusts and torques treated as quadratic with rotor speeds~\cite{hoffmann2007quadrotor,mahony2012multirotor,pounds2010modelling,hua2013introduction}. Aerodynamic effects such as vertical rotor inflow~\cite{pounds2007design,martin2010true}, wind disturbances~\cite{waslander2009wind,arain2014real}, vortex ring state~\cite{shetty2011small}, blade flapping\cite{pounds2010modelling}, translational lift and drag, and other unsteady effects may or may not be explicitly included.  Addressing these problems has been a corner stone of UAV research for the past two decades and even to the present these phenomena drive research in search of improved modelling and control solutions~\cite{simplicio2024robust,10556855,he2020quasi,lee2023adaptive,9677985,poggi2024optimal}.

\begin{figure}[!t]
	\centering
\includegraphics[width=\columnwidth]{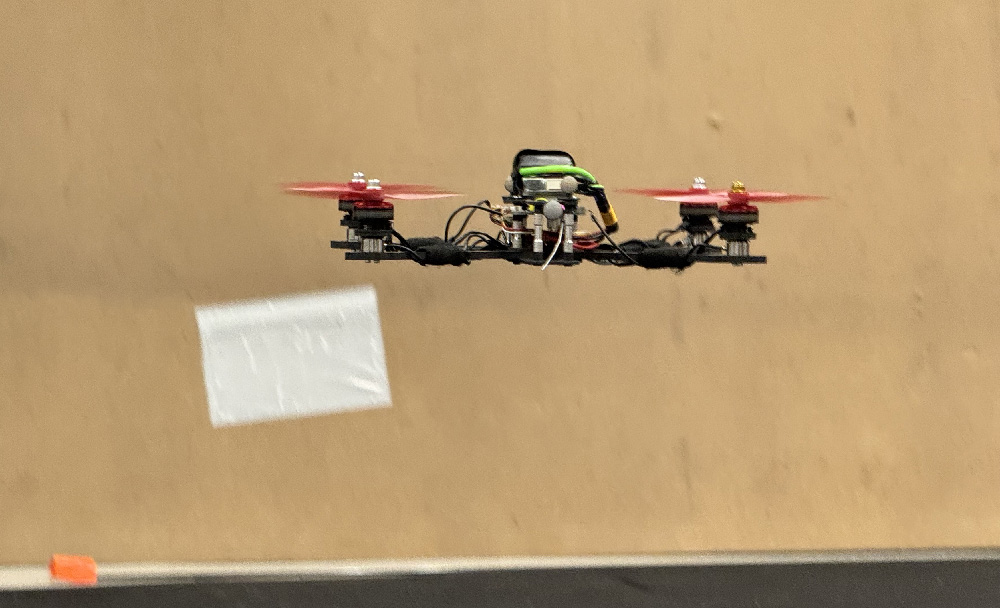}
	\caption{Quadrotor flying with integrated rotor force sensors.}
	\label{fig:flight}
    \vspace{-15px}
\end{figure}

Complex aerodynamic phenomena are challenging to model accurately, and simplified representations that do not capture the full gamut of non-stationary, extrinsic real world behaviors may lead to unanticipated divergence in flight~\cite{lupashin2010simple,kumar2017opportunities,de2018control}. Approaches to compensating for such phenomena in multirotors can be categorized into three areas: (i) improved detail to close the gap between the model and reality (ii) adaptive, learning approaches that refine an implicitly-learned model's parameters over time, and (iii) direct feedback control of the aerodynamic thrust\footnote{Simply ignoring unmodelled aerodynamics as disturbances and relying instead on empirical tuning and adjustment may also yield acceptable results.}.

Researchers have developed comprehensive models for rotors~\cite{martin2010true, bristeau2009role}, blade flapping~\cite{pounds2006acra} and translational lift~\cite{huang2009aerodynamics} and varying aerodynamic domains~\cite{orsag2009hybrid, powers2013influence,luo2015novel,abeywardena2013improved}. While extensive, these models are prescriptive and thus cannot capture unanticipated phenomena.  Conversely, ML approaches have been successful for aggressive maneuvers such as high-speed flight~\cite{kaufmann2023champion}, flips~\cite{lupashin2010simple}, and flying through narrow gaps and perching on inverted surfaces~\cite{mellinger2012trajectory}, but require training for specific maneuvers through repetition and iterative adjustments.  However, explicit modelling, learning and adaptive control based on vehicle dynamics alone are incapable of resolving disturbances before they can influence the aircraft's trajectory.

Our work considers the third approach, in which the output thrusts of the rotors are made to conform to that required to produce a desired trajectory. While this seems to be the case in all multirotor controllers, it is typically missing a measurement of the actual thrust generated by the rotors. In 2017 Bangura and Mahony used electrical power measurements to drive thrust estimators to compensate for translational lift and unmodeled axial and horizontal airflow disturbances~\cite{bangura2014aerodynamic,bangura2017thrust}, reducing tracking error compared to basic rotor speed control. A drawback of this approach is that changes in thrust are inferred from indirect upstream measurements, potentially limiting disturbance rejection fidelity.  In contrast, our approach is to use force transducer measurement of applied thrust at the rotor itself. Bangura and Mahony asserted that load cell sensors would be unsuccessful due to noise~\cite{bangura2017thrust}---a significant effort in this paper is therefore on testing this experimentally with a see-saw attitude control rig to better understand the feasibility of such an approach.

Direct thrust measurement for fixed-wing propellers using thin-film force sensors was explored in~\cite{bronz2017flight}---which showed significant noise due to sensor characteristics, vibrations and mechanical interactions---and strain gauges on a quadrotor frame for measuring rotor thrust are presented in~\cite{bazin2016feasibility}.  Both cases required inertial compensation, but neither employed thrust regulation; rather, the focus was on technical implementation of thrust measurement and demonstrating the feasibility of these solutions.  Beyond the authors' earlier 2016 work in drift compensation~\cite{davis2016sensor}, we are unaware of direct thrust measurements being used for multirotor thrust control purposes.

In this paper, we present a thrust control system for a multirotor see-saw model wherein the instantaneous force of the rotors is directly measured using load cells and controlled in a high-speed inner loop to regulate thrust.  Section~\ref{sec:model} describes the underlying philosophy of the approach, along with our dynamical model and control scheme.  Section~\ref{sec:hardware} presents the measurement, control and disturbance generating hardware driving our testbed.  Section~\ref{sec:unirotor} and~\ref{sec:see-saw} report experiments in controlling a single balancing rotor and `see-saw' test cases, respectively. Finally, we demonstrate that the added sensing and control elements do not preclude flight (see Fig.~\ref{fig:flight}) and provide benefit, and a brief conclusion completes the paper.

\section{Modeling and Control}\label{sec:model}
Classic multirotor PID control uses Inertial Measurement Unit (IMU) attitude error measurements and computes motor voltages to change rotor speed and induce desired forces.  Disturbances can enter this flow at two points: aerodynamic forces acting on the aircraft body, and air currents that change the inflow conditions of the rotor blades, altering the instantaneous rotor thrust~\cite{pounds2006acra}.

\begin{figure*}[!t]
	\centering
	\includegraphics[width=0.9\textwidth]{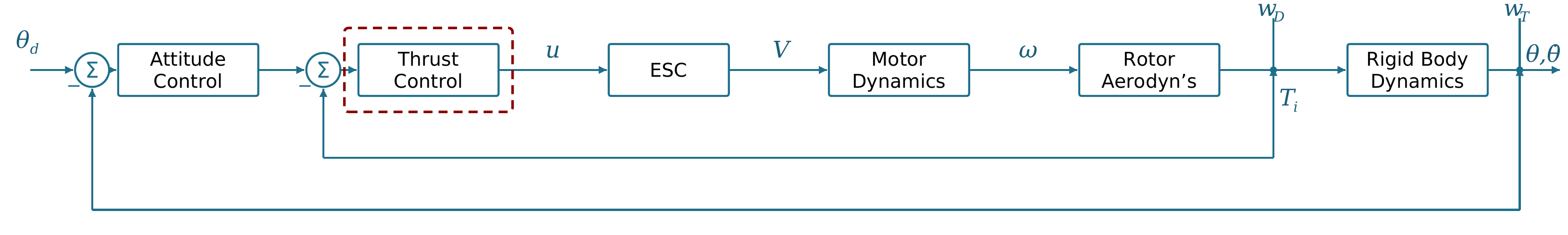}
	\caption{Quadrotor attitude control architecture with thrust regulation using two nested control loops. Thrust control can be bypassed with constant loop gain.}
	\label{fig:architecture}
    \vspace{-7.5px}
\end{figure*}

\subsection{Dynamical Model}
The basic planar dynamics of a quadrotor are given by
\begin{eqnarray}
    {m\ddot{x}}&=& -(T_{1} + T_{2})\sin\theta\\%
    {m\ddot{z}}&=& (T_{1} + T_{2})\cos\theta\\%
    {\textrm{I}\ddot{\theta}}&=& d_{1} T_{1} + d_{2} T_{2} + w%
\end{eqnarray}
where $x$ and $z$ are longitudinal and lateral position, $\theta$ is the pitch angle, $m$ and $\textrm{I}$ are the aircraft mass and rotational inertia, $d_{i}$ is the distance of the $i$th rotor axis from the Center of Gravity (CoG), $T_i$ is the thrust of the $i$th rotor, and $w$ is aerodynamic disturbance torque.

Instantaneous thrust is dependent on rotor speed, $\omega_{i}$, which in turn is driven by applied voltage $V_{i}$ and rotor drag torque, $Q_{i}$:
\begin{eqnarray}
T_{i} &=& k_{T}\omega_{i}^2 -k_{c}\omega \Delta v_{i}\\
Q_{i} &=& k_{Q}\omega_{i}^2
\end{eqnarray}
where $k_{T}$, $k_{Q}$ and $k_{c}$ are rotor thrust, drag and inflow damping coefficients, respectively~\cite{pounds2010modelling}.  $\Delta v_{i}$ is the rotor inflow change.

Changing inflow velocity at a rotor alters the local angle of attack of the blades, which changes the amount of lift produced; changing rotor inflow will also effect rotor drag torque, but this effect is small.  This variation in inflow velocity may be due to external gusts or the movement of the rotor through its own wake.  For movement through its own wake, the change in thrust is opposite the direction of motion, producing vertical inflow damping:
\begin{equation}
\Delta v_{i} = \dot{z} + d_{i}\dot{\theta} + v_{w}
\end{equation}
where $v_{w}$ is gust disturbance.

The dynamics of the rotor are given by:
\begin{equation}
\frac{\lambda}{R} V_{i} = \textrm{I}_{R}\dot{\omega_{i}} + \frac{\lambda^2}{R}\omega_{i} + k_{Q}\omega_{i}^2
\end{equation}
where $\textrm{I}_{R}$ is the rotor rotational inertia, $R$ is electrical resistance, and $\lambda$ is the motor flux linkage coefficient.

\subsection{Control Architecture}\label{sec:controller}
Aerodynamic effects induce variations in the forces acting on the vehicle that require adjustments in thrust to counteract these forces. We contextualise the control problem as a measurement problem rather than a purely analytical challenge: by utilizing direct high-speed control around thrust, we can more accurately follow the desired thrust trajectory. Measuring thrust at the point of generation avoids disturbances propagating into the rigid body dynamics which would require corrective action; thus, we expect the system robustness to disturbances to improve.

Our control architecture is a variation on classic nested-loop PID schemes, shown in Fig.~\ref{fig:architecture}. An outer PID loop regulates pitch angle by generating a thrust demand using the attitude error computed from the IMU. The inner loops regulate thrust to match the thrust references using error computed from the instantaneous rotor thrust measurements. Thrust is measured at the base of the motors using low-cost commodity load cells connected to a high-speed Analog to Digital Converter (ADC). This is then passed to a control module which generates signals to drive the motors. There is an independent thrust control loop for each rotor in the system.  The outer loop operates at \num{100}\si{\Hz}, which is the rate of the IMU updates. The inner loop operates at \num{500}\si{\Hz}, providing sufficient time to compensate for disturbances. The load cell measurements are read at \num{320}\si{\Hz}, the limit of the ADC, but the central loop is run faster to provide an over-sampled output to the motors.

In the multi-rotor setup, each rotor is equipped with its own load cell and a separate thrust measurement, resulting in multiple inner loops, one for each rotor. The set-points from the outer loop are complementary around zero.

In our experiments we compare our proposed thrust controller against a classical RPM controller; this same controller, with identical tuning parameters $K_p$, $K_d$, and $K_i$, serves as the outer loop (attitude controller) of the thrust feedback controller. This cascaded controller adds an inner loop that utilizes thrust measurements and outputs the Pulse-Width Modulation (PWM) signals for motor control. Both the classical controller and the outer loop controller aim to maintain the beam at zero degrees angle to the ground.  Meaningful comparison can be made to the typical open-loop thrust generation approach by bypassing the inner controller with a linear gain stage equal to the closed loop gain. The attitude controller is specifically set to be a poor-quality regulator so as to better demonstrate the influence of thrust regulation. Modern FCs use PID rate controllers and interating with and comparing with these will be future work.

\section{Test Rig and Disturbance Generation}
\label{sec:hardware}
\begin{figure}[t]
	\centering
	 \includegraphics[width=\columnwidth]{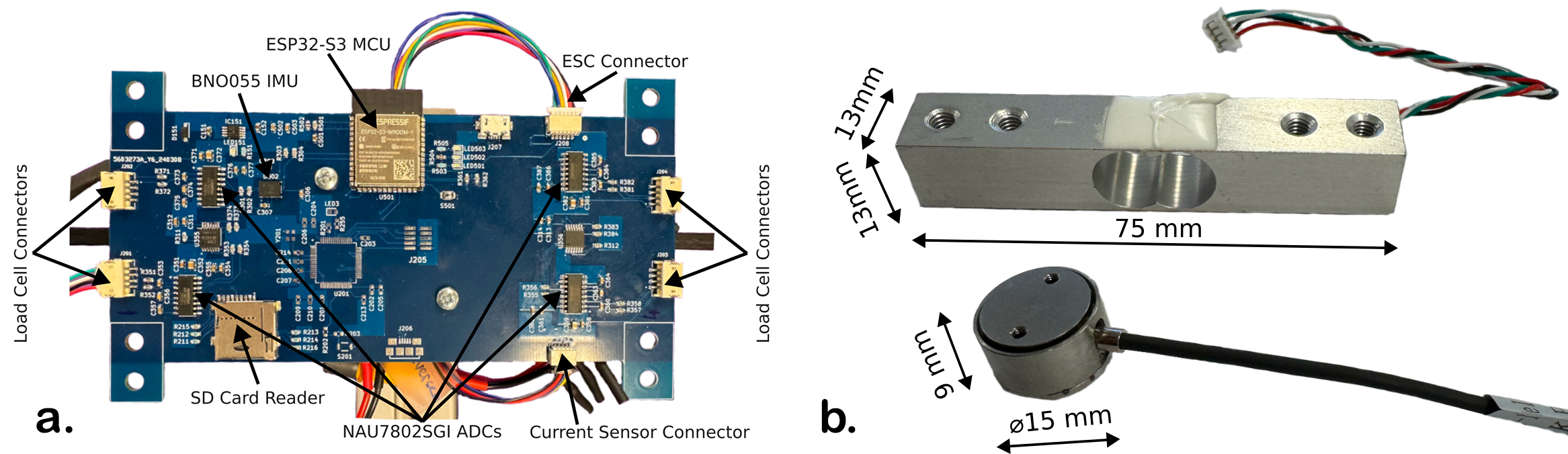}
	\caption{a. Custom PCB used in the experiments. It provides concurrent thrust measurements for 4 load cells at~\num{320}\si{\Hz}.  b. Load cells used in the experiments. Model numbers are CE08685 (beam) and DYZ-100 (column).}
	\label{fig:hardware}
    \vspace{-10px}
\end{figure}

\begin{figure}[t]
	\centering
     \includegraphics[width=\columnwidth]{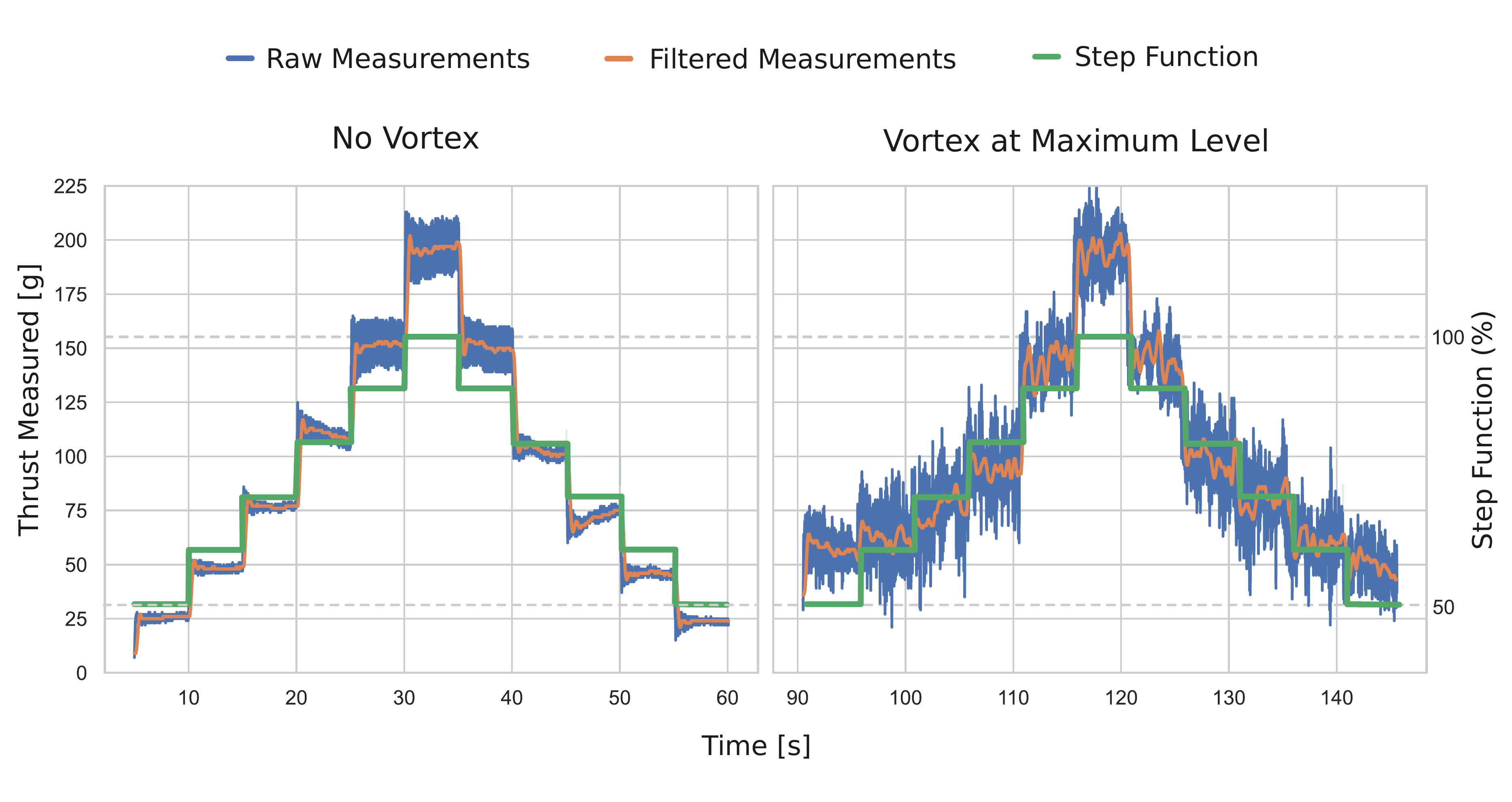}
	\caption{Effects of a low-pass Butterworth filter on raw thrust measurements. Data recorded onboard at \num{320}\si{\Hz}. The see-saw is fixed to remain parallel with the ground. The left side shows thrust measurements without turbulence, while the right side shows measurements during maximum turbulence. The stepping is controlled in an open-loop manner. The thrust between 90s and 100s is mostly generated by the turbulence.}
	\label{fig:thrust-filtering}
    \vspace{-10px}
\end{figure}

\begin{figure}[t]
	\centering
	\includegraphics[width=\columnwidth]{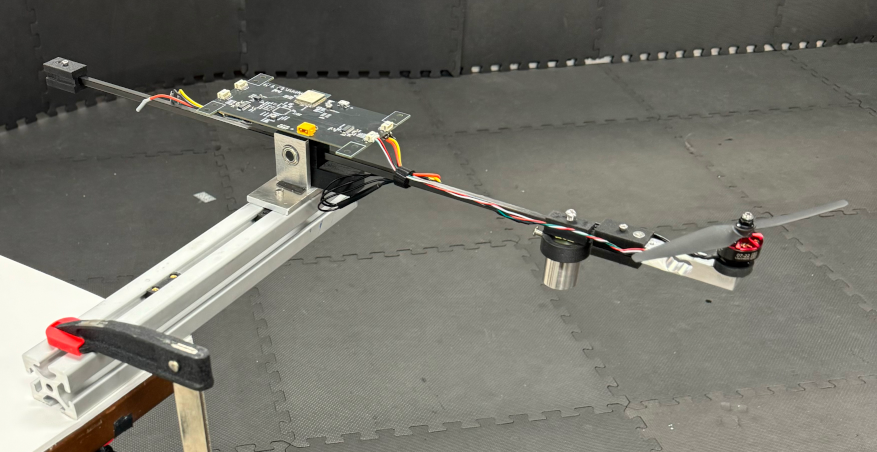}
	\caption{Single-rotor apparatus with rotor, balance weight, and load cell.}
	\label{fig:single-rotor_rig}
    \vspace{-10px}
\end{figure}

\begin{figure}[!t]
	\centering
	\includegraphics[width=\columnwidth]{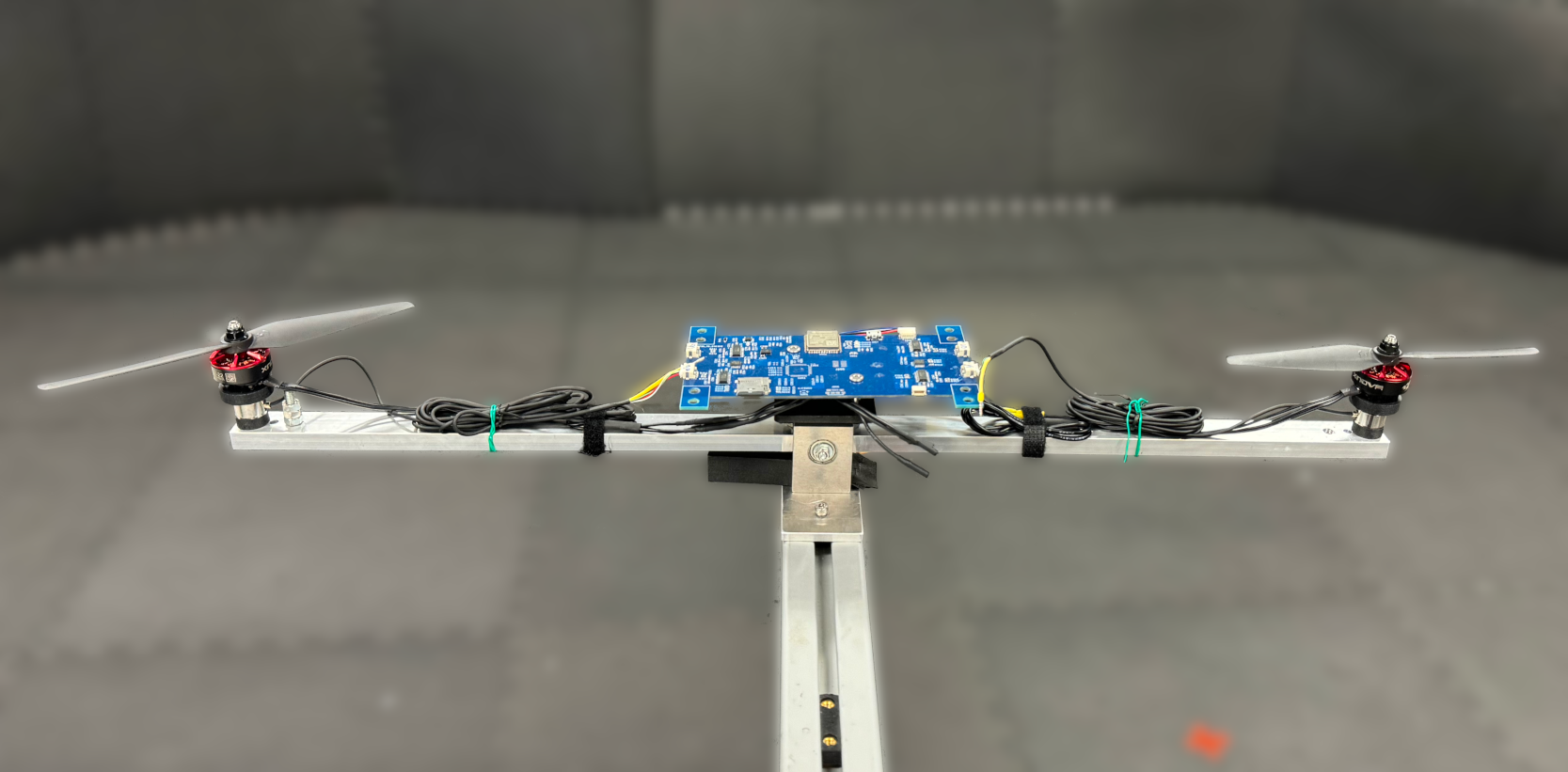}
	\caption{Dual-rotor apparatus with opposing thrusters and load cells.}
	\label{fig:see-saw}
    \vspace{-10px}
\end{figure}

Three experiments were performed to assess the effectiveness of direct thrust measurement and control: vertical inflow disturbance, crosswind disturbance, and transient ground effect. We conducted experiments using a see-saw apparatus that can be configured as either a single-rotor or dual-rotor setup. In single-rotor configuration (Figure~\ref{fig:single-rotor_rig}), only one side of the see-saw was actuated, counteracted by a \num{130}\si{\gram} fixed bias weight.  In the dual-rotor configuration, rotors were placed on both sides of the see-saw, as shown in Figure~\ref{fig:see-saw}.

\subsection{Sensors and Measurement}
Sensor readings are processed by an onboard controller running on an ESP32-S3-WROOM-1 MCU. Our custom quadrotor control PCB is shown in Figure~\ref{fig:hardware}a. 
The ESP32 generates PWM signals to control the motors through a four-in-one BLHELI-S ESC. Telemetry data is streamed using a WiFi connection at \num{100}\si{\Hz} using MQTT protocol for off-board processing; high frequency data is logged on-board using an SD card. Pitch angle measurements are provided by a BNO055 IMU, which uses onboard sensor fusion to generate an Euler vector at \num{100}\si{\Hz}.  A power board provides voltage and current measurements using INA226 current shunt and power monitor.

The thrust generated by the rotors is measured using commodity beam (model number CE08685) and column (model number DYZ-100) load cells rated for \num{1}\si{\kilogram}. Each motor is mounted directly on a load cell, and thrust measurements are sampled by an NAU7802SGI 24-bit Analog to Digital Converter (ADC) at~\num{320}\si{\Hz}. This ADC was chosen for its high sampling rate and compact form factor, allowing for onboard measurements from multiple load cells. This contrasts with the setup described by~\cite{bazin2016feasibility}, which required external cabling and connection to an off-board ADC unit (NI 9237).

We experimented with two types of low-cost load cells shown in Figure~\ref{fig:hardware}b. Although the higher-precision models provided less noisy measurements, they still required substantial filtering using a low-pass filter; best results were achieved with a Butterworth filter, using \num{320}\si{\Hz} as the sampling frequency and \num{2}\si{\Hz} as the cut-off frequency. The effects of this filtering are shown in Figure~\ref{fig:thrust-filtering}. This filter was used for all the experiments described in this paper.

\subsection{Vertical Inflow Disturbance Generation}
\begin{figure*}[t]
	\centering
    \includegraphics[width=0.80\textwidth]{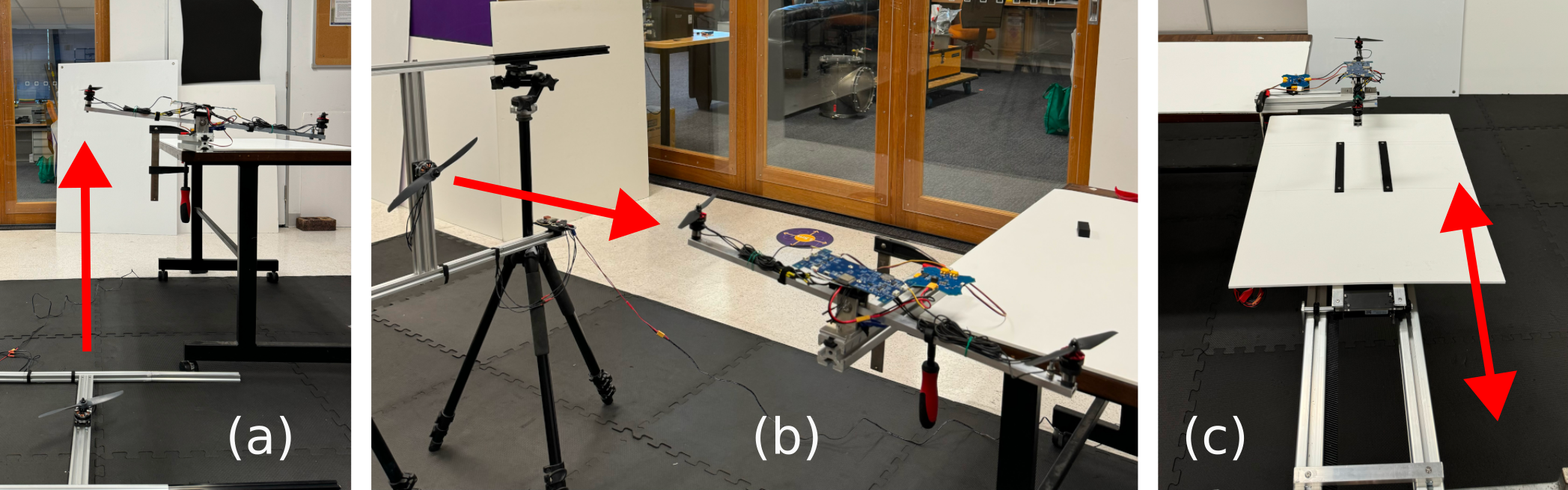}
	\caption{Dual-rotor experimental setup illustrating three experiments: (a) Vertical inflow experiment with upward-directed wind generated beneath the rotor, (b) crosswind experiment with wind generated parallel to the ground, and (c) ground effect experiment, where a plate is rapidly moved under the rotor and then retracted.}
    \vspace{-10px}
	\label{fig:dual-rotor_experiments}
\end{figure*}


We evaluate the system's response to vertical inflow disturbances, which are produced by a large rotor mounted directly beneath the apparatus and blowing upward, as shown in Figure~\ref{fig:dual-rotor_experiments}A. In each test, the rotor see-saw system is allowed to come to equilibrium and then the disturbance fan is engaged through an ascending and descending velocity step function.  The distance between the rotors is \num{70}\si{\cm}.

\begin{figure}[!t]
	\centering
 \includegraphics[width=\columnwidth]{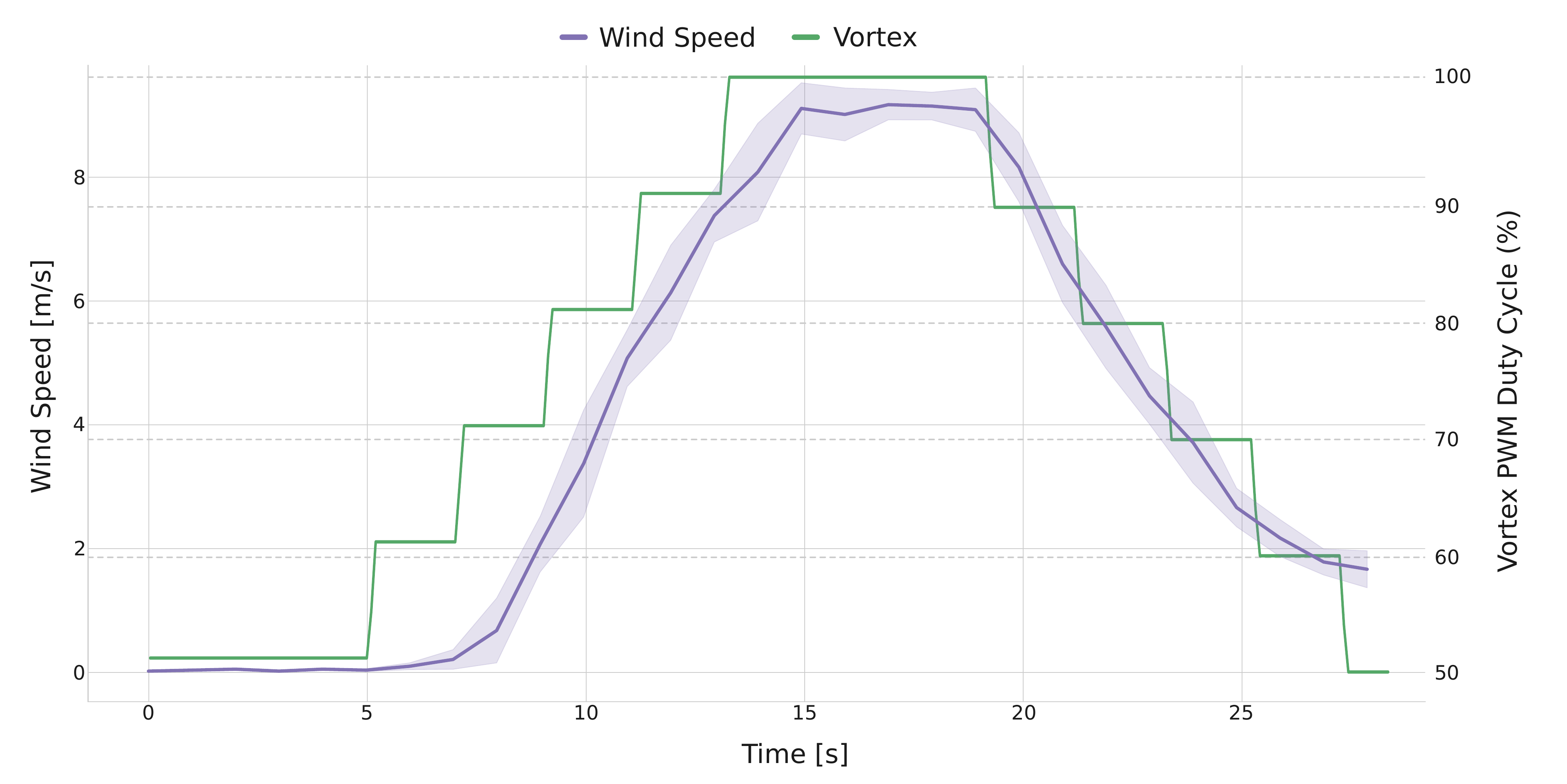}
	\caption{Disturbance rotor step function and the resulting wind speed. Anemometer readings averaged over 10 consecutive runs. Due to the anemometer's sparse measurement rate of one reading per second, fine details of the turbulence are not captured.}
	\label{fig:anemometer}
    \vspace{-10px}
\end{figure}

The disturbance rotor is controlled open-loop with throttle from \num{0}\si{\percent} (\num{0}\unit{m.s^{-1}}) to \num{100}\si{\percent} (\num{9.0}\unit{m.s^{-1}}). The approximate generated wind speeds during the disturbance sweep are illustrated in Figure~\ref{fig:anemometer}.

\begin{figure}[!t]
	\centering
	 \includegraphics[width=\columnwidth]{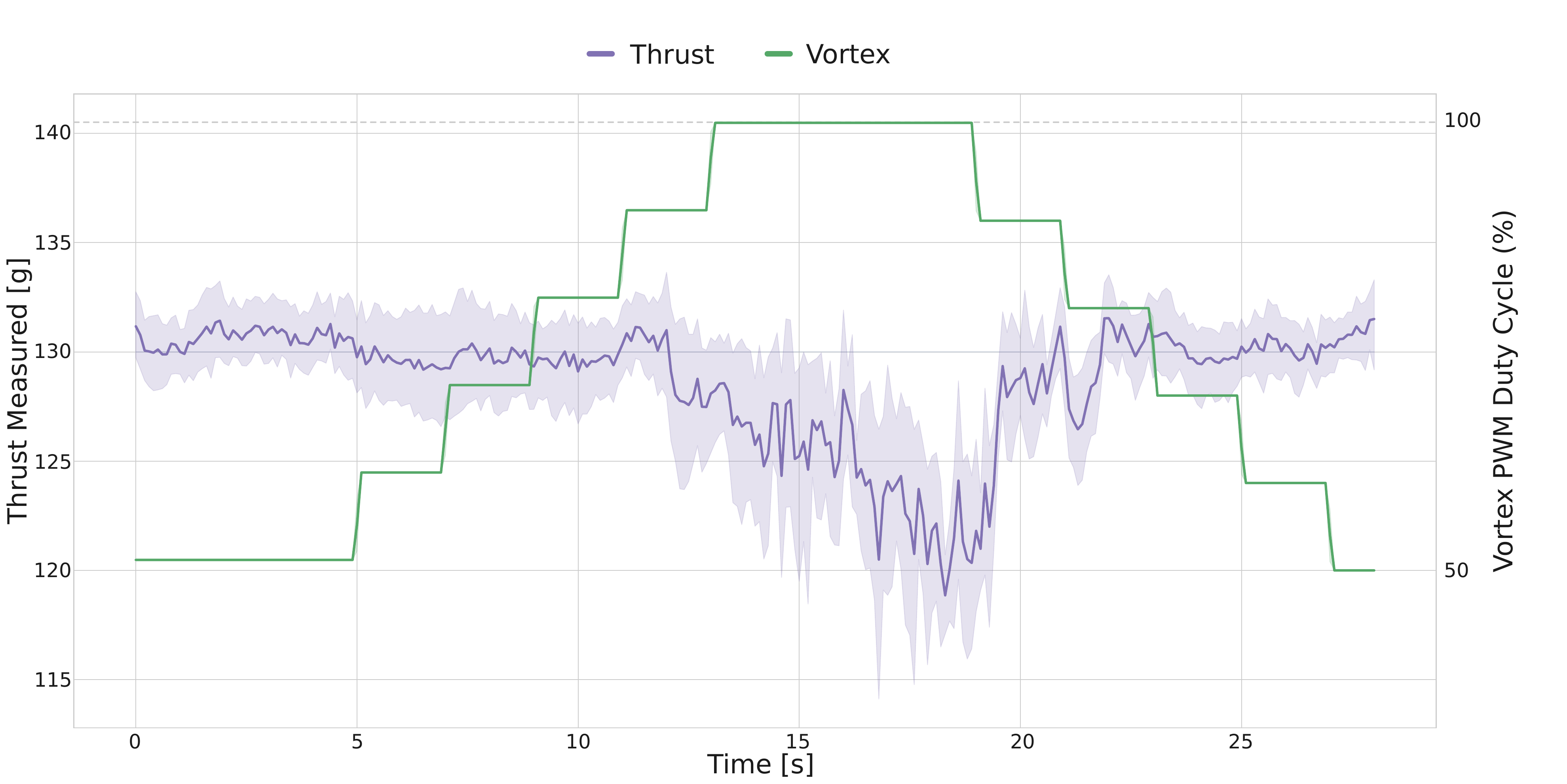}
	\caption{Effects of the disturbance on thrust measurement. The see-saw is fixed to remain parallel with the ground. The rotor operates with a fixed PWM to generate approximately the same thrust as the PID controller without disturbance. The green line shows the intensity of the disturbance. The drop in generated thrust indicates the presence of a vortex ring.}
	\label{fig:vortex-effects-open-loop}
    \vspace{-10px}
\end{figure}

The effects of the disturbance on the thrust measurements of a constant input, open loop fixed rotor in isolation are shown in Figure~\ref{fig:vortex-effects-open-loop}: the measured thrust decreases during maximum disturbance, indicating that the system is in the vortex ring state. Once the disturbance subsides and the system exits the vortex ring state, and returns to normal thrust output.


\subsection{Crosswind Disturbance Generation}
The crosswind was generated using the same large rotor used for vertical inflow, but positioned level with one of the test-rig rotors, \num{50}\si{\cm} in front of the apparatus, as shown in Figure~\ref{fig:dual-rotor_experiments}B. The magnitude of the wind was controlled using the same step function as per Figure~\ref{fig:anemometer}.


\subsection{Transient Ground Effect Generation}
Ground effect disturbance was generated using the apparatus shown in Figure~\ref{fig:dual-rotor_experiments}C. A \num{60}\si{\cm} $\times$ \num{50}\si{\cm} plate was moved under one of the rotors, positioned \num{7}\si{\cm} from the see-saw structure and \num{11}\si{cm} below the rotor.
The plate traveled a distance of \num{50}\si{cm}, moving from a point \num{30}\si{\cm} away from the rotor's center to \num{20}\si{\cm} inside past the center.
It then paused for \num{3}\si{s} before moving back out to its original position and remained stationary for another \num{3}\si{s}.
Each translation of the plate (in and out) took \num{1.6}\si{s}.

\section{Single-Rotor Thrust Balance}\label{sec:unirotor}
In a first experiment, we used as single-rotor rig.  We measured the single-rotor system's response to increasing and decreasing inflow disturbance, with and without thrust regulation, with results averaged over 10 consecutive runs --- operated continuously without any adjustments or interruptions.  We allowed \num{3}\si{\s} between the end of one disturbance and the start of the next to allow settling of air.

\begin{figure}[!h]
	\centering \includegraphics[width=\columnwidth]{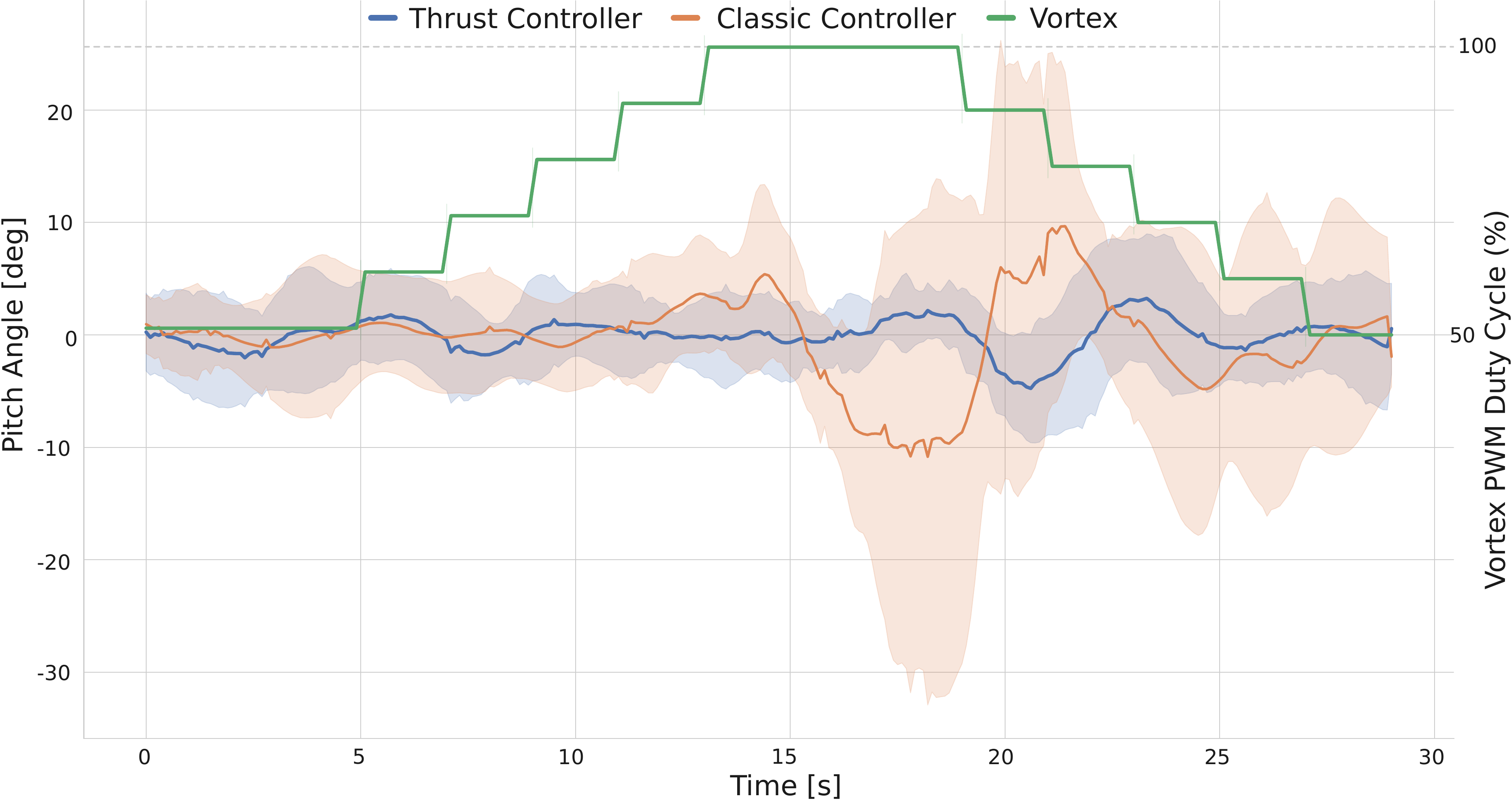}
	\caption{Single rotor balance during vertical inflow disturbance without thrust regulation, averaged over 10 consecutive runs. The disturbance rotor input function (not to scale) is overlaid for reference.}
	\label{fig:pitch-single-rotor}
\end{figure}

The classical balance controller without thrust regulation completely lost balance, with the see-saw reaching its deflection limit during each test.

In contrast, the thrust-regulated system was stable, with standard deviations shown in Figure~\ref{fig:pitch-single-rotor}. While both systems were affected by the disturbance, the thrust force controller maintained balance within $\pm$\num{10}\si{\degree}.

As shown in Figure~\ref{fig:vortex-effects-open-loop}, the disturbance reduces the efficiency of the rotor, manifested in a drop of generated thrust, for the same control signal. This causes the see-saw to drop at the start of the period of maximum disturbance (represented by a positive pitch angle in Figure~\ref{fig:pitch-single-rotor}).
This drop is compensated for by increased control action,
which then leads to a sudden upward movement as the disturbance reduces. This occurred during each of the 10 runs.

\begin{figure}
	\centering \includegraphics[width=\columnwidth]{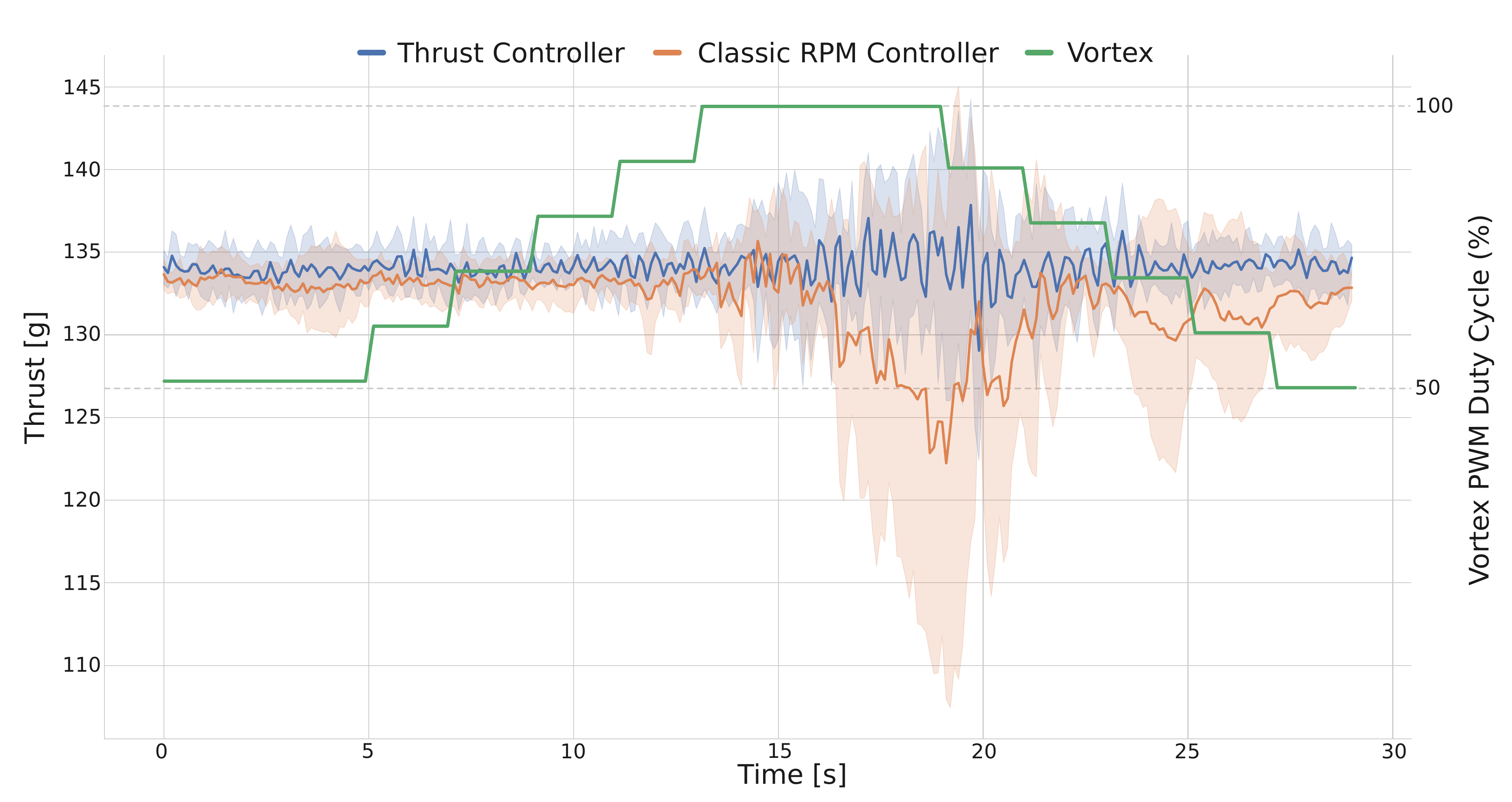}
	\caption{Single rotor balance with thrust regulation during vertical inflow disturbance of a pitch angle, averaged over 10 consecutive runs. The disturbance rotor input function (not to scale) is overlaid for reference.}
	\label{fig:thrust-single-rotor}
\end{figure}

Figure~\ref{fig:thrust-single-rotor} shows the thrust measurement recorded during the experiments.
The high inflow speeds lead to a vortex ring, which reduces the efficiency of the rotors at peak disturbance velocity, as evidenced in Figures~\ref{fig:vortex-effects-open-loop} and~\ref{fig:thrust-filtering} respectively. For the classical controller, this inefficiency manifests as an initial pitch drop (i.e., a positive angle) at the onset of maximum turbulence, due to reduced rotor efficiency. The pitch then recovers, slightly overshooting when the vortex fluctuates, and suddenly rises when the vortex diminishes, as rotor efficiency returns to normal. Conversely, the thrust controller maintains consistent thrust throughout and responds faster because it directly reacts to the measured changes in the generated thrust.


\section{Dual-Rotor `See-saw' Experiments}\label{sec:see-saw}
In the second set of experiments, a two-rotor rig was used with one side exposed to disturbance. Three trials were conducted, targeting a different aerodynamic disturbance: (i) vertical inflow, (ii) crosswind, and (iii) ground effect, respectively. As per the first experiment, the controllers were set to maintain orientation parallel to the ground by utilizing a zero-pitch set-point for the balance controller.

\subsection{Vertical Inflow Disturbance}
In this experiment, both rotors operated at a lower RPM compared to the single-rotor experiments, producing half the original thrust each (approximately \num{60}\si{\g}).  Otherwise, the protocol is identical to that of Section~\ref{sec:unirotor}.

\begin{figure*}[t]
	\centering \includegraphics[width=0.8\textwidth]{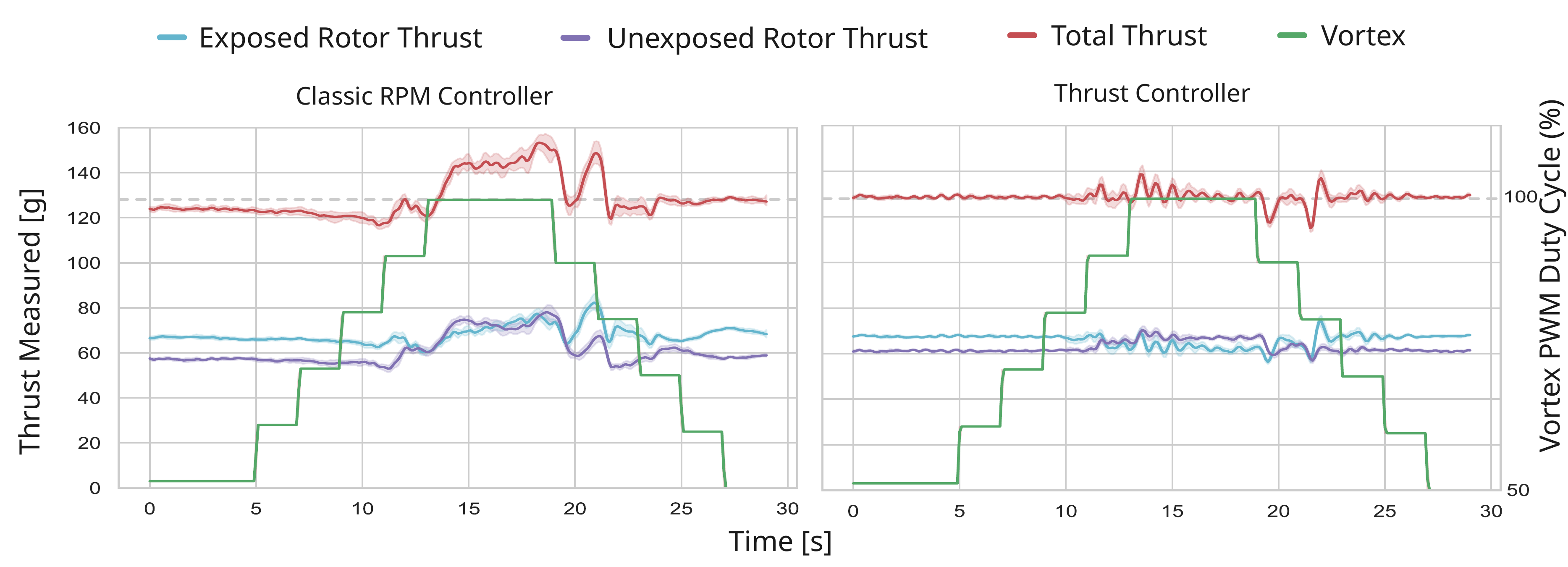}
	\caption{Thrust comparison in dual-rotor experiments. In the classic RPM controller, increased control signals during turbulence cause the net thrust to rise, potentially resulting in excess lift during flight. In contrast, the thrust controller maintains consistent overall thrust. }
	\label{fig:dual-rotor-thrust}
\end{figure*}

\begin{figure}
	\centering
    \includegraphics[width=\columnwidth]{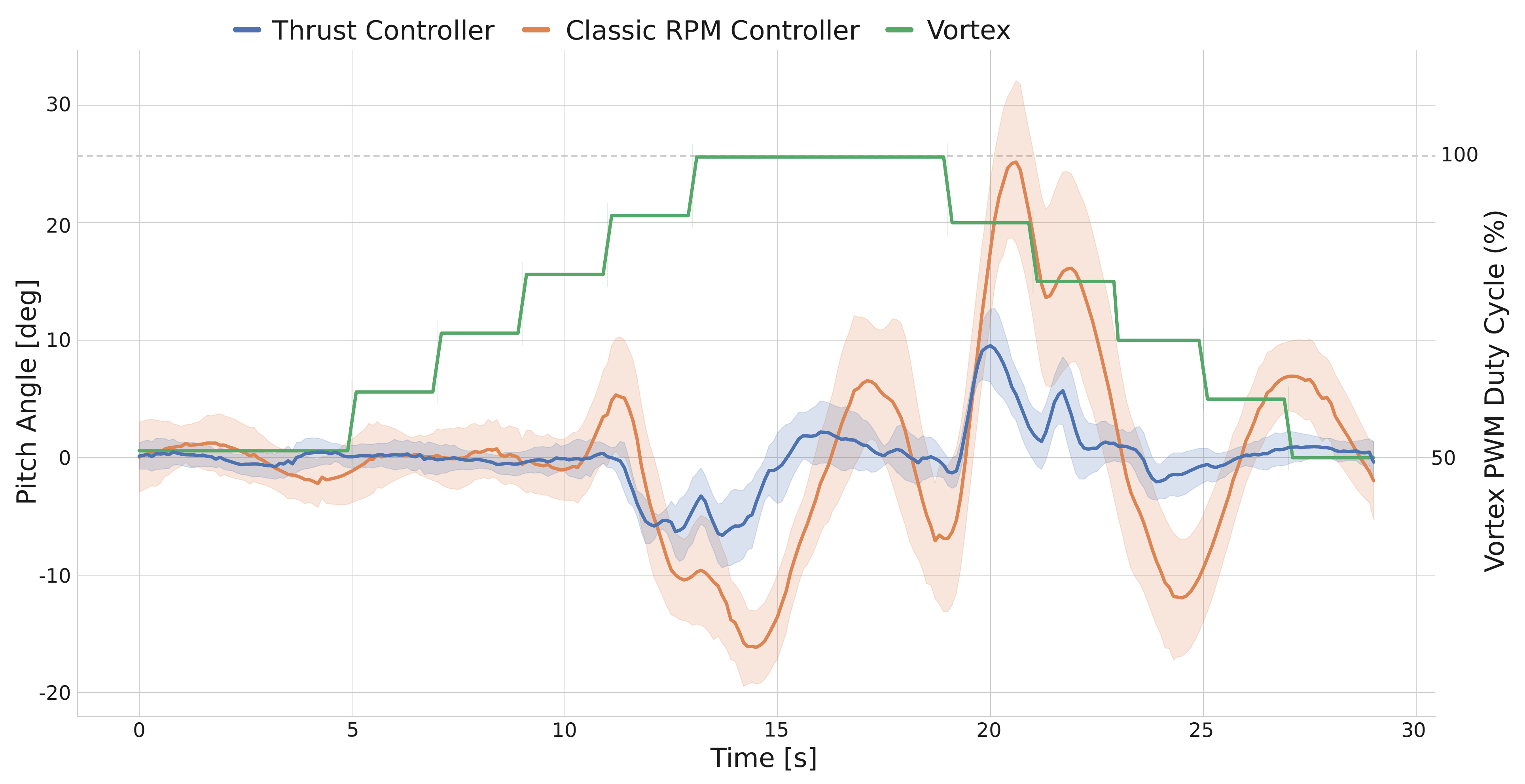}
	\caption{Pitch comparison in dual-rotor experiments. The classic RPM controller exhibits significantly larger pitch oscillations, consistently reaching maximum possible deflection. In contrast, the thrust controller maintains consistent pitch within $\pm$\num{10}\si{\degree}. }
	\label{fig:dual-rotor-pitch}
	\vspace{-10px}
\end{figure}

In the dual-rotor configuration, both controllers with and without thrust feedback are less affected by inflow disturbance compared to the single-rotor case, due to the inflow damping action of the second rotor.  However, there are significant differences between their pitch and thrust responses, illustrated in Figure~\ref{fig:dual-rotor-pitch} and in Figure~\ref{fig:dual-rotor-thrust}, respectively.

In the classical RPM controller without thrust feedback, the control signal is complementary around the throttle level, per basic quadrotor control allocation. This causes the control signal in the unexposed motor to increase, leading to an net increase in system thrust. Since the system is fixed at the pivot point, balance can be achieved at any thrust level, provided the thrust at each end of the apparatus is equal \textemdash~however, in free flight, this increase in overall thrust would result in unwanted lift.

The classical RPM controller loses control at the onset of the maximum disturbance and again when the disturbance input begins stepping down with a maximum deviation of $\pm$\num{30}\si{\degree}.  In contrast, the thrust feedback controller demonstrates significantly more stable pitch control, with thrust demand changing to suite the local flow environment at each rotor, limiting deviations to within $\pm$\num{10}\si{\degree}.

\subsection{Crosswind Disturbance Experiments}
When flying in  crosswinds, UAV rotors produce a nose-up pitching moment due to induced drag or blade flapping from variations in their local angles of attack and flow speeds relative to the rotating blades. Wind tunnel experiments conducted by Otsuka~\cite{otsuka2018reduction} at wind speeds similar to our own demonstrated that this pitching moment disrupts the control of the UAV's pitch, resulting in less stable posture in windy conditions compared to hovering in calm air.

\begin{figure}
	\centering
    \includegraphics[width=1.0\columnwidth]{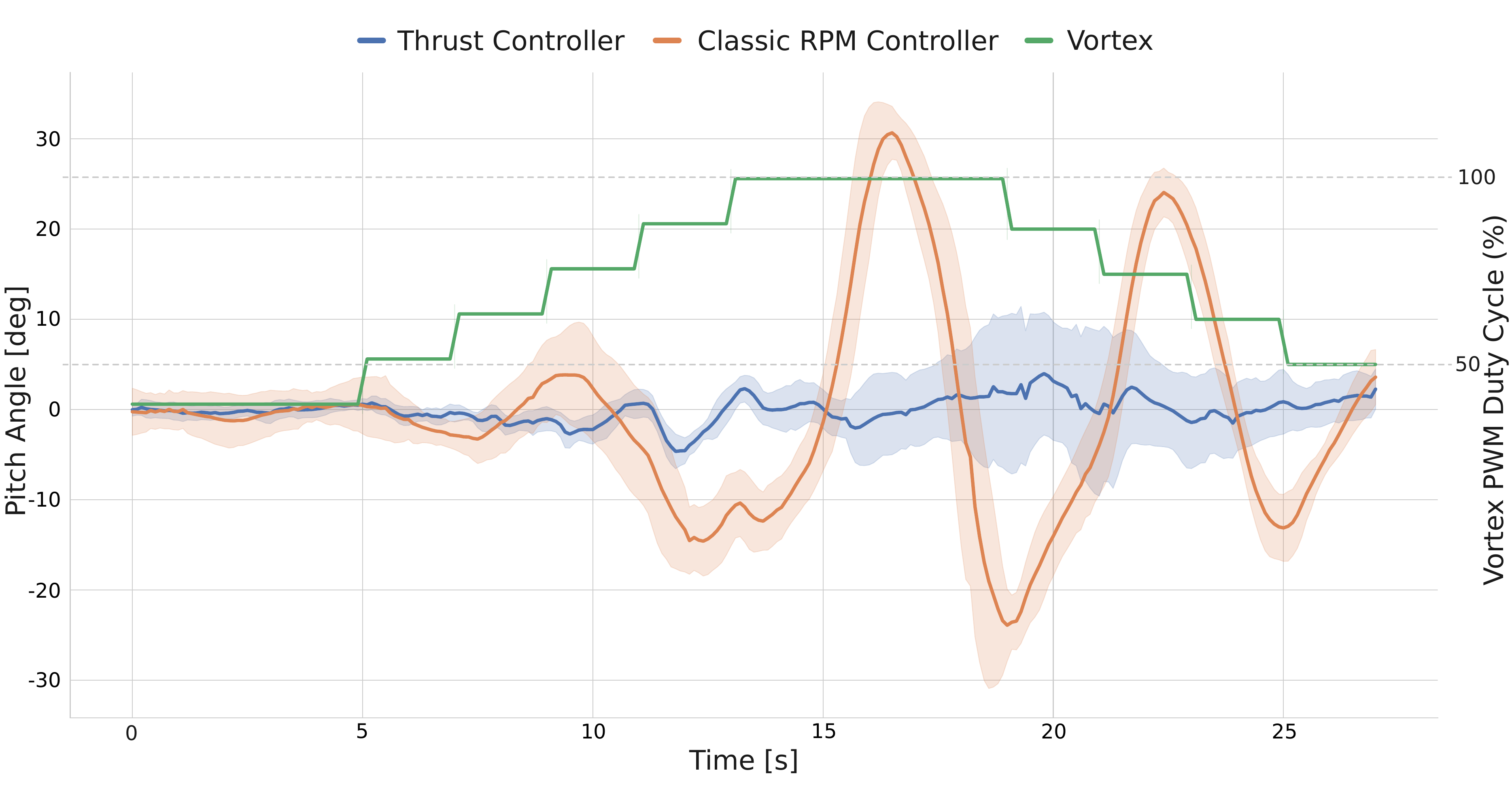}
	\caption{Pitch comparison in dual-rotor experiments with crosswind. The classic RPM controller experiences substantial pitch oscillations, frequently reaching the maximum deflection limits. By comparison, the thrust controller maintains a stable pitch, keeping deviations within $\pm$\num{10}\si{\degree}.}
	\label{fig:pitch-comparison-headwind}
\end{figure}

It was found that the classic controller exhibited significant oscillations, with the system reaching both the minimum and maximum deflection limits, per Figure~\ref{fig:pitch-comparison-headwind}. An upward pitching moment was observed between 10 and 15 seconds, commencing when the wind speed reached approximately \num{6}\unit{m.s^{-1}} and intensifying as the wind speed increased. The controller attempted to compensate for this upward movement by increasing thrust in the second rotor; however, the strong wind caused overshooting, pushing the system to its maximum negative deflection.

In contrast, the thrust controller maintained stable pitch within $\pm$\num{10}\si{\degree}, effectively mitigating the wind-induced disturbances under the same conditions.

\subsection{Ground Effect Disturbance Experiments}
Ground effect refers to the increased lift and decreased aerodynamic drag experienced by an aircraft when it operates close to a fixed surface, such as the ground (typically within one rotor diameter of altitude).  In our experiment, we subjected one rotor to intermittent ground effect conditions, as described in Section\ref{sec:hardware}.  This sequence was repeated ten times.

\begin{figure}
	\centering
	 \includegraphics[width=1.0\columnwidth]{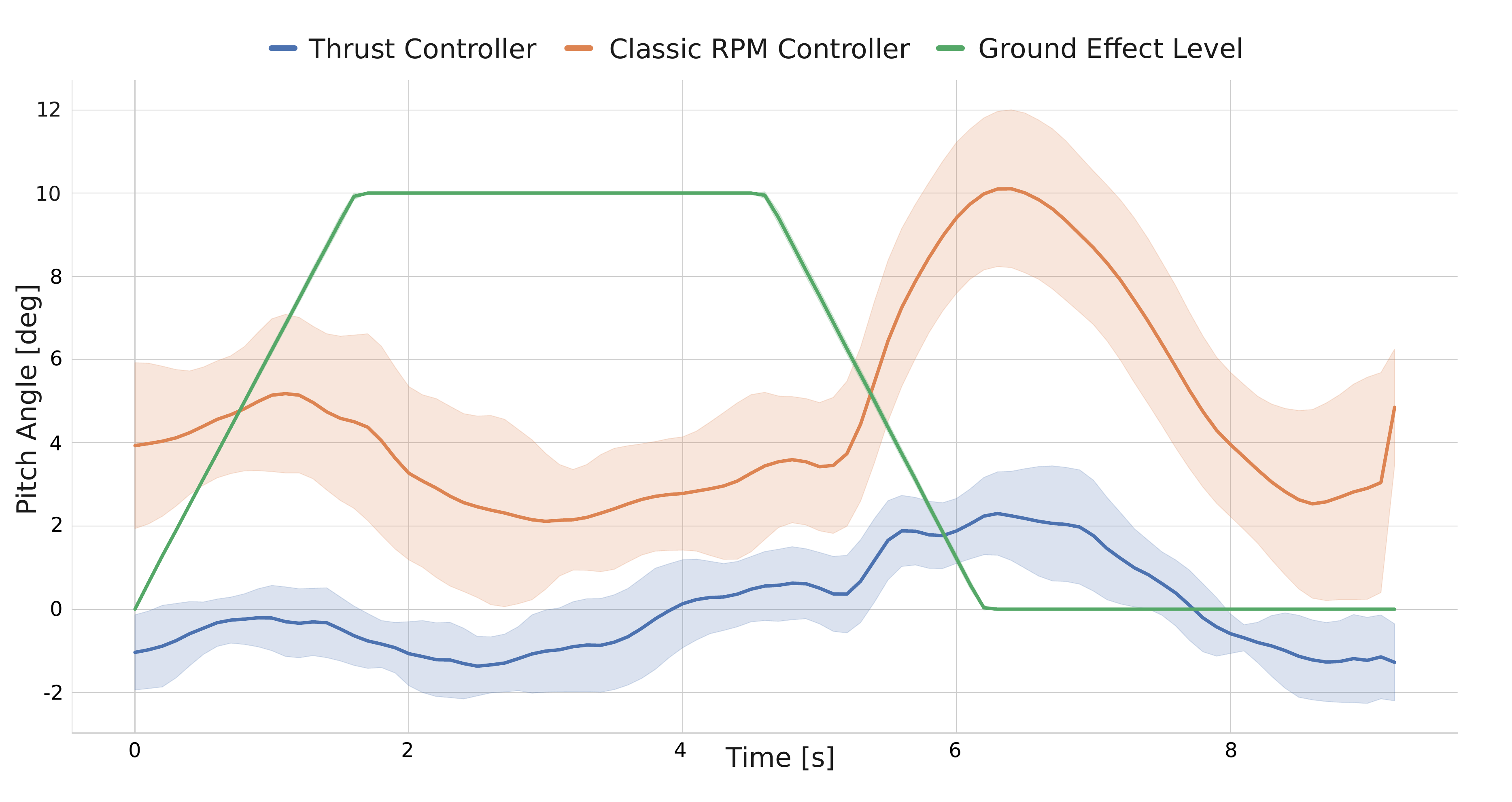}
	\caption{Pitch comparison in dual-rotor experiments with ground effect. While the classic RPM controller exhibits significant pitch drops --- exceeding \num{10}\si{\degree} --- when the ground plane moves under and then away from one of the rotors, the thrust controller maintains a stable pitch with only minor deviations within \num{3}\si{\degree}.}
	\label{fig:pitch-comparison-ground-effect}
	\vspace{-10px}
\end{figure}

Our results, summarized in Figure~\ref{fig:pitch-comparison-ground-effect}, illustrate this effect. The increased lift caused by ground effect led the classic controller to exhibit a pitch deflection when the plate moved in, followed by a more significant overshoot of approximately \num{10}\si{\degree} when the plate withdrew. In contrast, the thrust force controller maintained stable pitch control, experiencing only a minor deflection of approximately \num{2}\si{\degree} when the plate withdrew.

The control techniques described above were employed with the addition of a feed-forward term to provide a throttle set-point not relying on integral action, and addition of extra filtering.  The filters comprised both digital and analog stages to reduce noise and allow for higher proportional control gain suited to fast drone dynamics. Insufficient filtering at sufficiently high levels induced vibration.

\section{Preliminary Flight Test Validation}\label{sec:flight_validation}
On the basis of the positive results reported above, we constructed a test quadrotor platform capable of flying with both force feedback control active or bypassed, using the same attitude control time constants for each.  The quadrotor used  \mbox{DYZ-100} load cells with integrated filtering to reduce signal noise: see Fig.~\ref{fig:flight}.

In manually directed free flight, the aircraft's subjective handling characteristics were found to be better under thrust control, described by the pilot as `less floaty' and easier to direct, compared to the conventional performance of the classic controller.  To provide metric comparison to validate that the benefit was not idiosyncratic to the pilot, we applied a PID position controller to the aircraft and directed it to maintain a fixed position within a test cell instrumented with Optitrack Prime13 motion tracking.

\begin{figure*}[b]
	\centering \includegraphics[width=\textwidth]{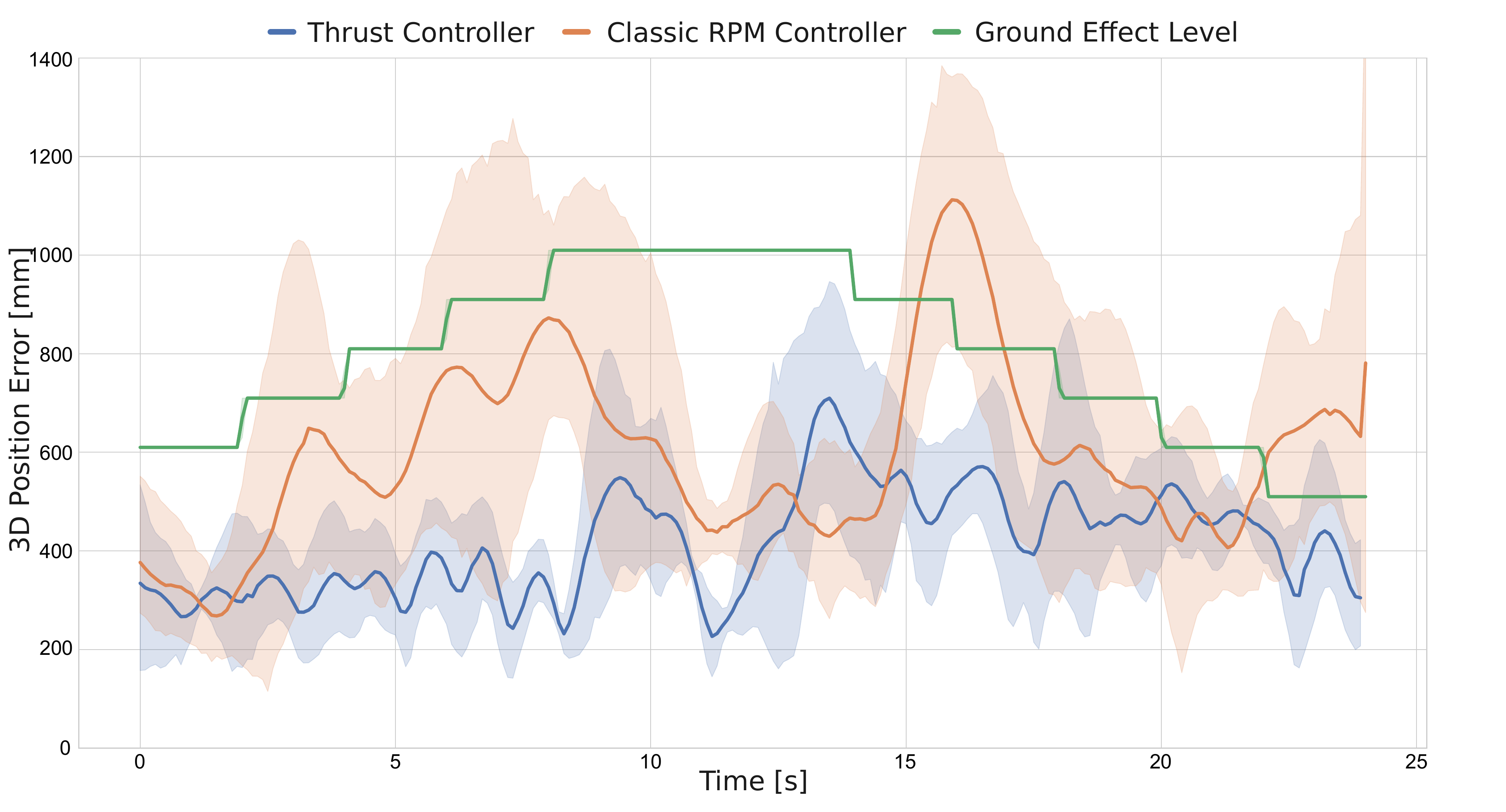}
	\caption{Proof-of-concept free flight over the vertical disturbance fan with and without thrust feedback control, respectively.}
	\label{fig:flight-experiments}
\end{figure*}

During flight, a sequence of increasing upward wind disturbance velocities were applied from the same fan used for the vertical inflow disturbance experiments, above. We repeated the disturbance steps 10 times each for both thrust control and classic control architectures, and then computed the normalized 3D position error magnitude run-average to assess how well each controller responded to the applied disturbance.  Results of the experiment are shown in Figure~\ref{fig:flight-experiments}.

It was found that the thrust control architecture had uniformly lower normalized 3D position error than the classic controller.  The magnitude of error peaked strongly for the classic controller during the onset and end of thrust disturbance, indicating that it was less capable of adjusting to changing aerodynamic conditions than our thrust-regulating controller.

On the basis of this result, we are confident that the thrust control concept is able to integrate with flying platforms and provide a genuine benefit in flight handling.  Our future work will extensively test this free-flying implementation under the same range of disturbance conditions used for the see-saw to verify that the full suite of anticipated performance improvements are achieved in flight.

\section{Conclusion}
\label{sec:conclusion}
This paper introduced a rotor control system using fast feedback of direct rotor force measurement using load cells for thrust regulation under disturbance.  This thrust control system was applied for both a single rotor on a pivot stand and a multi-rotor see-saw model, using purpose-built hardware from low-cost commodity components.  The thrust controllers demonstrated both the feasibility of this approach in the presence of complex aerodynamic phenomena, and were compared against an attitude PID controller producing RPM demands, which we refer to as a \("\)classical RPM controller\("\) with identical same tuning gains.

Experiments demonstrated robustness to transverse disturbances, vertical inflow disturbances, and applied ground effect, respectively.  The proposed controller with high-speed force feedback regulation successfully regulated attitude in the face of these disturbances and outperformed the RPM-based controller in every case, substantially reducing the influence of the disturbances.

We demonstrated a basic proof-of-concept system showing that the technique can be implemented in flying hardware.  This demonstrate showed subjectively better flying qualities for pilots and enhanced ability to weather changing aerodynamic disturbances.

This work shows great potential to enable multirotor robot platforms to easily reject aerodynamic disturbances without substantial investment in modelling or adaptive trajectory tuning.  Future work will refine the quadrotor UAV platform to explore sensors and modalities for further improvement, and test the concept across a full range of common disturbances.

\section*{Acknowledgments}
The authors would like to acknowledge William Deer's contributions to this project.

\bibliographystyle{IEEEtran}
\bibliography{bibfile}

\vfill

\end{document}